\documentclass{article}

\usepackage{times}
\usepackage{graphicx}
\usepackage{subfigure}
\usepackage{capt-of}
\usepackage{natbib}
\usepackage[ruled,algo2e,linesnumbered]{algorithm2e}
\usepackage{todonotes}
\usepackage{multirow}
\usepackage{amsmath,amsfonts,amsthm,amssymb}
\usepackage{symbols}
\usepackage{placeins}

\usepackage{hyperref}

\usepackage[arxiv]{icml2015}

\icmltitlerunning{Generative Moment Matching Networks}

\begin{document} 

\twocolumn[
\icmltitle{Generative Moment Matching Networks}

\icmlauthor{Yujia Li${}^1$}{yujiali@cs.toronto.edu}
\icmlauthor{Kevin Swersky${}^1$}{kswersky@cs.toronto.edu}
\icmlauthor{Richard Zemel${}^{1,2}$}{zemel@cs.toronto.edu}
\icmladdress{${}^1$Department of Computer Science, University of Toronto, Toronto,
ON, CANADA\\${}^2$Canadian Institute for Advanced Research, Toronto, ON,
CANADA}

%\icmlauthor{Your CoAuthor's Name}{email@coauthordomain.edu}
%\icmladdress{Their Fantastic Institute,
%            27182 Exp St., Toronto, ON M6H 2T1 CANADA}

\icmlkeywords{deep learning, mmd, neural networks, generative models, graphical models}

\vskip 0.3in
]

\begin{abstract}
We consider the problem of learning deep generative models from data.
We formulate a method that generates an independent sample via a single
feedforward pass through a multilayer preceptron, as in the recently
proposed generative adversarial networks \cite{goodfellow2014generative}. Training a
generative adversarial network, however, requires careful optimization of a difficult
minimax program.  Instead, we utilize a technique
%Our approach is
%a variation of the recently proposed generative adversarial network
%\cite{goodfellow2014generative}. These networks
%have the appealing property that generating an independent sample requires a single
%feed-forward pass through a multilayer perceptron. Training a generative adversarial
%network, however, requires careful optimization of a difficult minimiax program.
%We propose instead to use a technique 
from statistical hypothesis testing known
as maximum mean discrepancy (MMD), 
%. This gives
which leads to a simple objective that can be interpreted as matching all orders of statistics
between a dataset and samples from the model, and can be trained
by backpropagation.
We further boost the performance of this
approach by combining our generative network with an auto-encoder network, using MMD
to learn to generate codes that can then be decoded to produce samples. We show
that the combination of these techniques yields excellent generative models compared to baseline approaches as measured on MNIST and the Toronto Face Database.
\end{abstract} 

\section{Introduction}
The most visible successes in the area of deep learning have come from the
application of deep models to supervised learning tasks. Models such as
convolutional neural networks (CNNs), and long short term memory (LSTM)
networks are now achieving impressive results on a number of tasks such as
object recognition \cite{krizhevsky2012imagenet,sermanet2014overfeat,szegedy2014going}, speech
recognition \cite{graves2014towards,deepSpeechReviewSPM2012}, image caption generation
\cite{vinyals2014show,fang2014captions,kiros2014unifying}, machine translation
\cite{cho2014learning,sutskever2014sequence},
and more. Despite their successes, one of the main bottlenecks of the
supervised approach is the difficulty in obtaining enough data to learn
abstract features that capture the rich structure of the data. It is well
recognized that a promising avenue is to use unsupervised learning on
unlabelled data, which is far more plentiful and cheaper to obtain. 

A long-standing and inherent problem in unsupervised
learning is defining a good method for evaluation. Generative models offer
the ability to evaluate generalization in the data space, which
can also be qualitatively assessed. 
In this work we propose a generative model for unsupervised learning that we
call generative moment matching networks (GMMNs). GMMNs are generative neural
networks that begin with a simple prior from which it is easy to draw samples.
These are propagated deterministically through the hidden layers of the network and the output
is a sample from the model. Thus, with GMMNs it is easy to quickly draw
independent random samples, as opposed to expensive MCMC procedures that are
necessary in other models such as Boltzmann machines
\cite{ackley1985learning,hinton2002training,salakhutdinov2009deep}. The
structure of a GMMN is most analogous to the recently proposed generative
adversarial networks (GANs) \cite{goodfellow2014generative}, however unlike GANs, whose training involves a difficult minimax optimization problem, GMMNs are comparatively simple; they are trained to minimize a straightforward loss function using backpropagation.

The key idea behind GMMNs is the use of a statistical hypothesis testing
framework called maximum mean discrepancy~\cite{gretton2007kernel}. Training a GMMN to minimize this discrepancy can be interpreted as matching all moments of the model distribution to the empirical data distribution. Using the kernel trick, MMD can be represented as a simple loss function that we use as the core training objective for GMMNs. Using minibatch stochastic gradient descent, training can be kept efficient, even with large datasets.

As a second contribution, we show how GMMNs can be used to bootstrap
auto-encoder networks in order to further improve the generative process.
%We call these networks generative moment matching auto-encoders (GMMN+AE). 
The idea behind this approach is to train an auto-encoder network and then apply a
GMMN to the code space of the auto-encoder. This allows us to leverage the
rich representations learned by auto-encoder models as the basis for comparing
data and model distributions. To generate samples in the original data space,
we simply sample a code from the GMMN and then use the decoder of the
auto-encoder network.

Our experiments show that this relatively simple, yet very flexible framework
is effective at producing good generative models in an efficient manner. On
MNIST and the Toronto Face Dataset (TFD) we demonstrate improved results over
comparable baselines, including GANs. Source code for training GMMNs will be
made available at \url{https://github.com/yujiali/gmmn}.

\section{Maximum Mean Discrepancy}
Suppose we are given two sets of samples $X = \{ x_i \}_{i=1}^N$ and $Y = \{
y_j \}_{j=1}^M$ and are asked whether the generating distributions $P_X =
P_Y$. Maximum mean discrepancy is a frequentist estimator for answering this
question, also known as the two sample test
\cite{gretton2007kernel,gretton2012kernel}. The idea is simple: compare statistics between the two datasets and if they are similar then the samples are likely to come from the same distribution.

Formally, the following MMD measure computes the mean squared difference of
the statistics of the two sets of samples.
\vspace{-5pt}
\begin{align}
    &\mathcal{L}_{\mathrm{MMD}^2} = \left \|\frac{1}{N} \sum_{i=1}^N \phi(x_i) -
    \frac{1}{M} \sum_{j=1}^M \phi(y_j)\right \|^2 \\
&= \frac{1}{N^2} \sum_{i=1}^N \sum_{i'=1}^N \phi(x_i)^\top \phi(x_{i'}) -
    \frac{2}{NM}\sum_{i=1}^N \sum_{j=1}^M \phi(x_i)^\top \phi(y_j) \nonumber\\
&+ \frac{1}{M^2}\sum_{j=1}^M \sum_{j'=1}^M \phi(y_j)^\top \phi(y_{j'}) \label{eq:mmdprimal}
\end{align}
Taking $\phi$ to be the identity function leads to matching the sample mean,
and other choices of $\phi$ can be used to match higher order moments.

Written in this form, each term in Equation~(\ref{eq:mmdprimal}) only involves inner products
between the $\phi$ vectors, and therefore the kernel trick can be applied.
\begin{align}
\mathcal{L}_{\mathrm{MMD}^2} &= \frac{1}{N^2} \sum_{i=1}^N \sum_{i'=1}^N k(x_i,
    x_{i'}) - \frac{2}{NM}\sum_{i=1}^N \sum_{j=1}^M k(x_i, y_j) \nonumber\\
&+ \frac{1}{M^2}\sum_{j=1}^M \sum_{j'=1}^M k(y_j, y_{j'}) \label{eq:mmddual}
\end{align}
The kernel trick implicitly lifts the sample vectors into an infinite
dimensional feature space. When this feature space corresponds to a universal
reproducing kernel Hilbert space, it is shown
that asymptotically, MMD is 0 if and only if
$P_X=P_Y$~\cite{gretton2007kernel,gretton2012kernel}. 

For universal kernels like the Gaussian kernel, defined as
$k(x,x')=\exp(-\frac{1}{2\sigma}|x-x'|^2)$, where $\sigma$ is the bandwidth
parameter, we can use a Taylor expansion to get
an explicit feature map $\phi$ that contains an infinite number of terms and
covers all orders of statistics. Minimizing MMD under this feature expansion
is then equivalent to minimizing a distance between \emph{all} moments of the
two distributions.

\section{Related Work}
In this work we focus on generative models due to their ability to capture the
salient properties and structure of data. Deep generative models are
particularly appealing because they are capable of learning a latent manifold
on which the data has high density. Learning this manifold allows smooth
variations in the latent space to result in non-trivial transformations in the
original space, effectively traversing between high density modes through low
density areas \cite{bengio2013better}. They are also capable of disentangling
factors of variation, which means that each latent variable can become
responsible for modelling a single, complex transformation in the original
space that would otherwise involve many variables \cite{bengio2013better}.
Even if we restrict ourselves to the field of deep learning, there are a vast array of approaches to generative modelling. Below, we outline some of these methods.

One popular class of generative models used in deep learning are undirected
graphical models, such as Boltzmann machines \cite{ackley1985learning},
restricted Boltzmann machines \cite{hinton2002training}, and deep Boltzmann
machines \cite{salakhutdinov2009deep}. These models are normalized by a typically intractable partition function, making training, evaluation, and sampling extremely difficult, usually requiring expensive Markov-chain Monte Carlo (MCMC) procedures.

Next there is the class of fully visible directed models such as fully visible sigmoid belief networks
\cite{neal1992connectionist} and
the neural autoregressive distribution estimator \cite{larochelle2011neural}. These admit efficient log-likelihood calculation, gradient-based learning and efficient sampling, but require that an ordering be imposed on the observable variables, which can be unnatural for domains such as images and cannot take advantage of parallel computing methods due to their sequential nature.

More related to our own work, there is a line of research devoted to recovering density models from auto-encoder networks using MCMC procedures \cite{rifai2012generative,bengio2013generalized,bengio2014deep}. These attempt to use contraction operators, or denoising criteria in order to generate a Markov chain by repeated perturbations during the encoding phase, followed by decoding.

Also related to our own work, there is the class of deep, variational networks~\cite{rezende2014stochastic,kingma2014variational,mnih2014belief}. These are also deep, directed generative models, however they make use of an additional neural network that is designed to approximate the posterior over the latent variables. Training is carried out via a variational lower bound on the log-likelihood of the model distribution. These models are trained using stochastic gradient descent, however they either require that the latent representation is continuous~\cite{kingma2014variational}, or require many secondary networks to sufficiently reduce the variance of gradient estimates in order to produce a sufficiently good learning signal~\cite{mnih2014belief}.

Finally there is some early work that proposed the idea of using
feed-forward neural networks to learn generative models.
\citet{mackay1995bayesian} proposed a model that is closely related to ours,
which also used a feed-forward network to map the prior samples to the data
space. However, instead of directly outputing samples, an extra distribution
is associated with the output. Sampling was used extensively for learning and inference in
this model. \citet{magdon1998neural} proposed to use a neural network to learn
a transformation from the data space to another space where the transformed
data points are uniformly distributed. This transformation network then learns the cumulative density
function.

\section{Generative Moment Matching Networks}

\subsection{Data Space Networks}
% data space net
The high-level idea of the GMMN is to use a neural network to 
learn a deterministic mapping from samples of a simple, easy to sample distribution,
to samples from the data distribution. 
The architecture of the generative network is exactly the same as a
generative adversarial network \cite{goodfellow2014generative}. However, we
propose to train the network by simply minimizing the MMD criterion, avoiding
the hard minimax objective function used in generative adversarial network
training.

More specifically, in the generative network we have a stochastic hidden layer
$\hv\in\real^H$ with $H$ hidden units at the top
with a prior uniform distribution on each unit independently,
\begin{equation}
p(\hv) = \prod_{j=1}^H U(h_j)
\end{equation}
Here $U(h)=\frac{1}{2}\Iv[-1\le h\le 1]$ is a uniform distribution in
$[-1,1]$, where $\Iv[.]$ is an indicator function.  Other choices for the
prior are also possible, as long as it is a simple enough distribution from which we
can easily draw samples.

\begin{figure}[t]
    \centering
    \begin{tabular}{cc}
        \includegraphics[width=0.38\columnwidth]{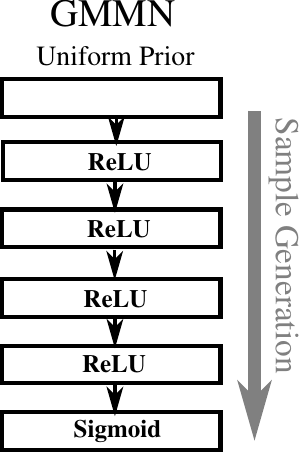} &
        \includegraphics[width=0.55\columnwidth]{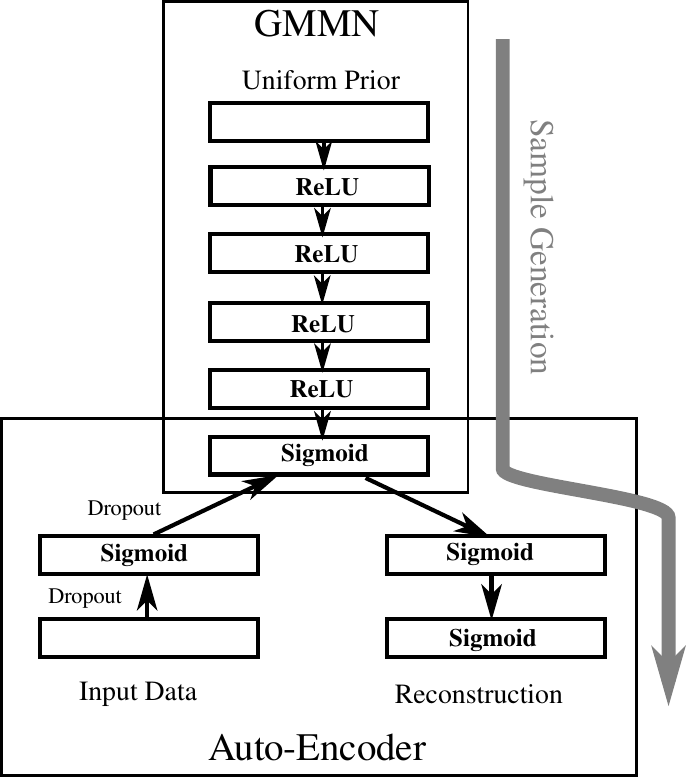} \\
        (a) GMMN & (b) GMMN+AE
    \end{tabular}
    \vspace{-5pt}
    \caption{Example architectures of our generative moment matching networks. (a) GMMN used in the input data space. (b) GMMN used in the code
    space of an auto-encoder.}
    \vspace{-5pt}
    \label{fig:architecture}
\end{figure}

The $\hv$ vector is then passed through the neural network and deterministically mapped
to a vector $\xv\in\real^D$ in the $D$ dimensional data space.
\begin{equation}
\xv = f(\hv;\wv)
\end{equation}
$f$ is the neural network mapping function, which can contain multiple layers
of nonlinearities, and $\wv$ represents the parameters of the neural network.
One example architecture for $f$ is illustrated in
Figure \ref{fig:architecture}(a), which has 3 intermediate ReLU
\cite{nair2010rectified} nonlinear layers and one logistic sigmoid output layer.

The prior $p(\hv)$ and the mapping $f(\hv;\wv)$ jointly defines a distribution
$p(\xv)$ in the data space. To generate a sample $\xv\sim p(\xv)$ we only need
to sample from the uniform prior $p(\hv)$ and then pass the sample $\hv$
through the neural net to get $\xv=f(\hv;\wv)$.

% If $f$ is invertible, by the change of variable technique we have
% \begin{equation}
% p(\hv) = p(\xv) \left|\frac{\mathrm{d}\xv}{\mathrm{d} \hv}\right|
% \end{equation}
% where $\frac{\mathrm{d}\xv}{\mathrm{d}\hv}=\nabla f(\hv;\wv)$ is the Jaccobian
% matrix of $f$. Therefore we have
% \begin{equation}
% p(\xv) = \frac{1}{|\nabla f(\hv;\wv)|} p(\hv)
% \end{equation}
% However, usually $f$ is not guaranteed to be invertible when modeled by a
% neural network.  But the distribution can still be represented by the samples
% generated from it. 

\citet{goodfellow2014generative} proposed to train this
network by using an extra discriminative network, which tries to distinguish
between model samples and data samples.  The generative network is then trained to counteract this in order to make the samples indistinguishable to the discriminative network.  The gradient of this
objective can be backpropagated through the generative network. However,
because of the minimax nature of the formulation, it is easy to get stuck at a
local optima. So the training of generative network and the discriminative network must be interleaved and carefully scheduled.  By contrast, our learning algorithm simply involves minimizing the MMD objective.

Assume we have a
dataset of training examples $\xv_1^d, ..., \xv_N^d$ ($d$ for data), and a set of samples
generated from our model $\xv_1^s, ..., \xv_M^s$ ($s$ for samples).
The MMD objective $\mathcal{L}_{\MMD^2}$ is differentiable when the kernel is differentiable.  For
example for Gaussian kernels $k(\xv,\yv) = \exp\left(-\frac{1}{2\sigma}||\xv -
\yv||^2\right)$, the gradient of $\xv^s_{ip}$ has a simple form
\begin{align}
    \pdiff{\mathcal{L}_{\MMD^2}}{\xv^s_{ip}} =& \frac{2}{M^2} \sum_{j=1}^M \frac{1}{\sigma}k(\xv^s_i,
    \xv^s_j)(\xv^s_{jp} - \xv^s_{ip}) \nonumber \\
    &- \frac{2}{MN} \sum_{j=1}^N
    \frac{1}{\sigma} k(\xv^s_i, \xv^d_j)(\xv^d_{jp} - \xv^s_{ip})
\end{align}
This gradient can then be backpropagated through the generative network to
update the parameters $\wv$.

\subsection{Auto-Encoder Code Space Networks}
Real-world data can be complicated and high-dimensional, which is one reason why generative modelling is such a difficult task. Auto-encoders, on the other hand, are designed to solve an arguably simpler task of reconstruction. If trained properly, auto-encoder models can be very good at representing data in a code space that captures enough statistical information that the data can be reliably reconstructed.

The code space of an auto-encoder has several advantages for creating a
generative model. The first is that the dimensionality can be explicitly
controlled. Visual data, for example, while represented in a high dimension
often exists on a low-dimensional manifold. This is beneficial for a
statistical estimator like MMD because the amount of data required to produce
a reliable estimator grows with the dimensionality of the data
\cite{ramadas2015decreasing}. The second
advantage is that each dimension of the code space can end up representing
complex variations in the original data space. This concept is referred to in the
literature as disentangling factors of variation \cite{bengio2013better}.

For these reasons, we propose to bootstrap auto-encoder models with a GMMN to
create what we refer to as the GMMN+AE model. These operate by first learning
an auto-encoder and producing code representations of the data, then freezing
the auto-encoder weights and learning a GMMN to minimize MMD between generated codes and data codes. A visualization of this model is given in Figure \ref{fig:architecture}(b).

Our method for training a GMMN+AE proceeds as follows:
\begin{enumerate}
    \item Greedy layer-wise pretraining of the auto-encoder
        \cite{bengio2007greedy}.
\item Fine-tune the auto-encoder.
\item Train a GMMN to model the code layer distribution using an MMD objective on the final encoding layer.
\end{enumerate}

We found that adding dropout to the encoding layers can be beneficial in terms
of creating a smooth manifold in code space. This is analogous to the
motivation behind contractive and denoising auto-encoders
\cite{rifai2011contractive,vincent2008extracting}.

\subsection{Practical Considerations}
Here we outline some design choices that we have found to improve the peformance of GMMNs.

\textbf{Bandwidth Parameter.} The bandwidth parameter in the kernel plays a
crucial role in determining the statistical efficiency of MMD, and optimally
setting it is an open problem. A good heuristic is to perform a line search to
obtain the bandwidth that produces the maximal distance~\cite{sriperumbudur2009kernel}, other more advanced
heuristics are also available
\cite{gretton2012optimal}. As a simpler
approximation, for most of our experiments we use a mixture of $K$ kernels
spanning multiple ranges. That is, we choose the kernel to be:
\begin{align}
k(x,x') &= \sum_{q=1}^K k_{\sigma_q}(x,x')
\end{align}
where $k_{\sigma_q}$ is a Gaussian kernel with bandwidth parameter $\sigma_q$.
We found that choosing simple values for these such as $1$, $5$, $10$, etc.
and using a mixture of 5 or more was sufficient to obtain good results. The
weighting of different kernels can be further tuned to achieve better results,
but we kept them equally weighted for simplicity.

\textbf{Square Root Loss.} In practice, we have found that better results can be obtained by optimizing
$\mathcal{L}_{\MMD} = \sqrt{\mathcal{L}_{\MMD^2}}$. This loss can
be important for driving the difference between the two distributions as close to 0 as
possible. Compared to $\mcL_{\MMD^2}$ which flattens out when its value gets close to 0,
$\mcL_\MMD$ behaves much better for small $\mcL_\MMD$ values. Alternatively, 
this can be understood by writing down the gradient of $\mcL_\MMD$ with respect to $\wv$
\begin{equation}
\pdiff{\mathcal{L}_\MMD}{\wv} = \frac{1}{2\sqrt{\mathcal{L}_{\MMD^2}}}
\pdiff{\mathcal{L}_{\MMD^2}}{\wv}
\end{equation}
The $1/(2\sqrt{\mathcal{L}_{\MMD^2}})$ term automatically adapts the
effective learning rate. This is especially beneficial when
both $\mathcal{L}_{\MMD^2}$ and $\pdiff{\mathcal{L}_{\MMD^2}}{\wv}$ become small,
where this extra factor can help by maintaining larger gradients. 

\textbf{Minibatch Training.} One of the issues with MMD is that the usage of kernels means that the
computation of the objective scales quadratically with the amount of data. In
the literature there have been several alternative estimators designed to overcome
this \cite{gretton2012kernel}. In our case, we found that it was sufficient to
optimize MMD using minibatch optimization. In each weight update, a small subset of
data is chosen, and an equal number of samples are drawn from the GMMN. Within
a minibatch, MMD is applied as usual. As we are using exact samples from the
model and the data distribution, the minibatch MMD is still a good estimator of
the population MMD. We found this approach to be both fast and effective. The
minibatch training algorithm for GMMN is shown in Algorithm \ref{alg:gmmn}.

\begin{algorithm2e}[t]
%\DontPrintSemicolon
\SetEndCharOfAlgoLine{}
\SetKwInOut{Input}{Input}
\SetKwInOut{Output}{Output}
\Input{Dataset $\{\xv^d_1, ..., \xv^d_N\}$, prior $p(\hv)$, network
$f(\hv;\wv)$ with initial parameter $\wv^{(0)}$}
\Output{Learned parameter $\wv^*$}
\BlankLine
\While{Stopping criterion not met}{
Get a minibatch of data $\Xv^d \gets\{\xv^d_{i_1}, ...,
\xv^d_{i_b}\}$\;
Get a new set of samples $\Xv^s\gets \{\xv^s_1, ...,
\xv^s_b\}$\;
Compute gradient $\pdiff{\mathcal{L}_\MMD}{\wv}$ on $\Xv^d$ and $\Xv^s$\;
Take a gradient step to update $\wv$\;
}
\caption{GMMN minibatch training}
\label{alg:gmmn}
\end{algorithm2e}
\vspace{-5pt}

\section{Experiments}
\begin{figure*}[t!]
    \centering
    \begin{tabular}{ccc}
        \multirow{2}{*}[3.5em]{\includegraphics[width=0.25\textwidth]{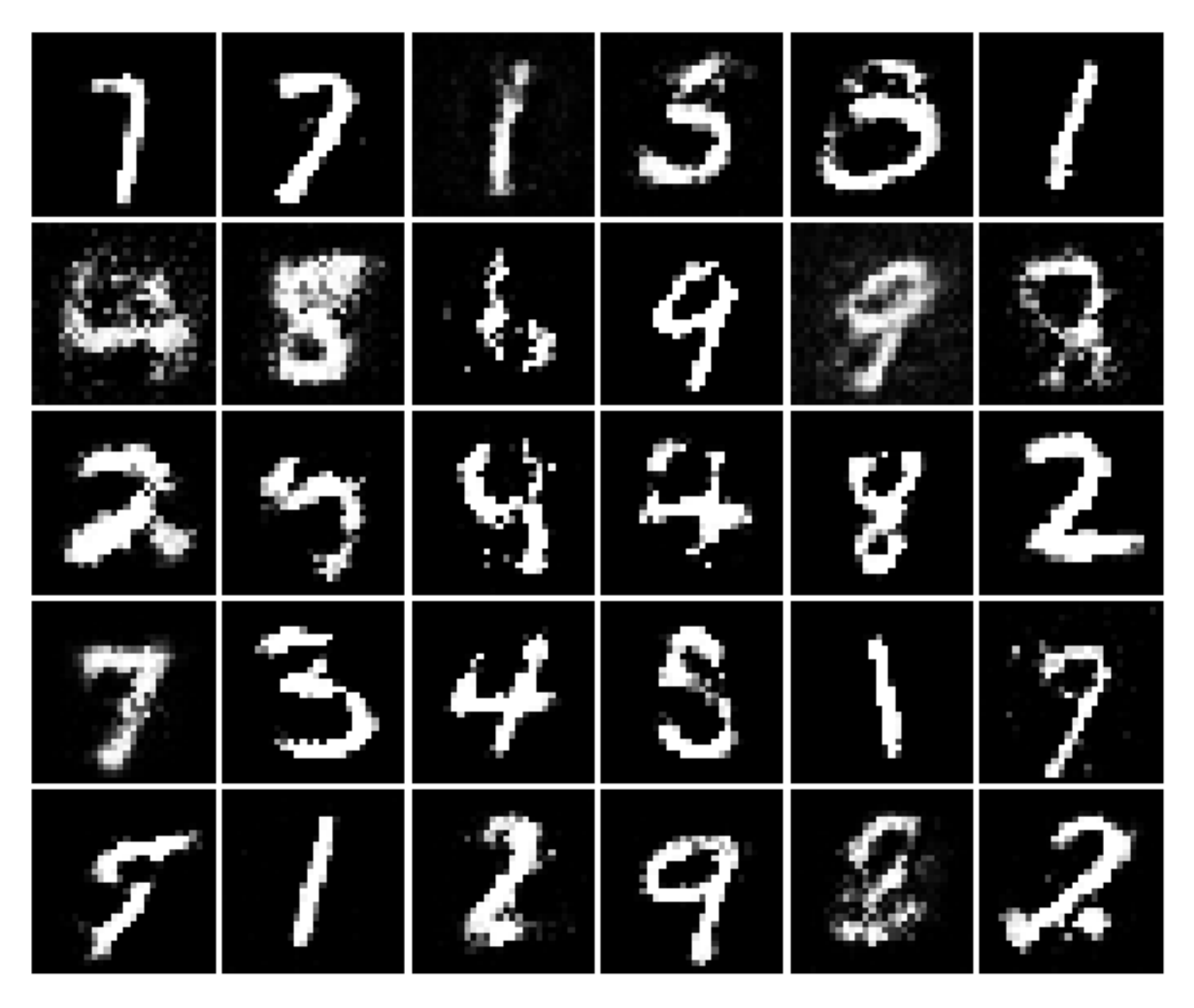}}
        &
        \multirow{2}{*}[3.5em]{\includegraphics[width=0.25\textwidth]{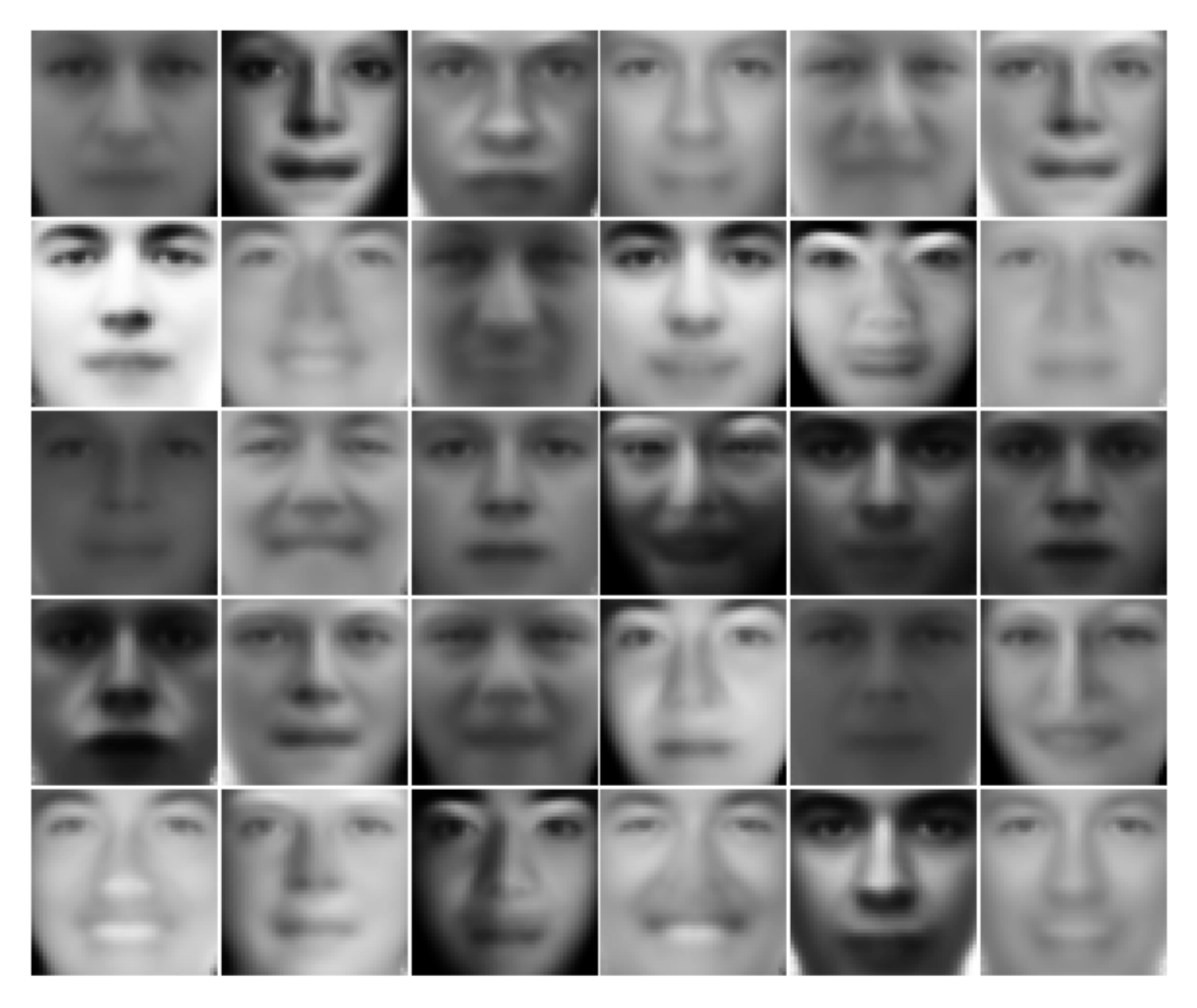}}
        &
        \includegraphics[width=0.45\textwidth]{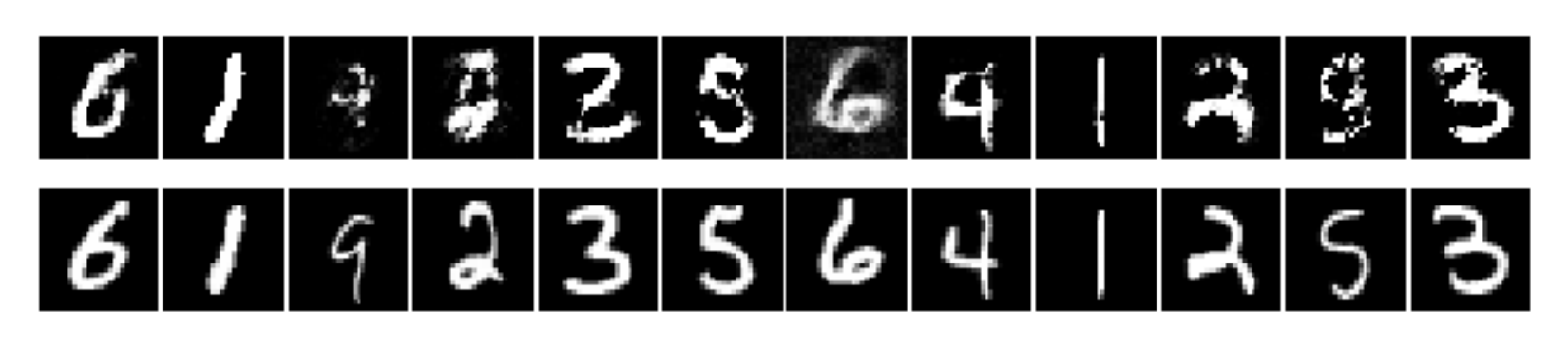} \\
        & & (e) GMMN nearest neighbors for MNIST samples \\
        & & \includegraphics[width=0.45\textwidth]{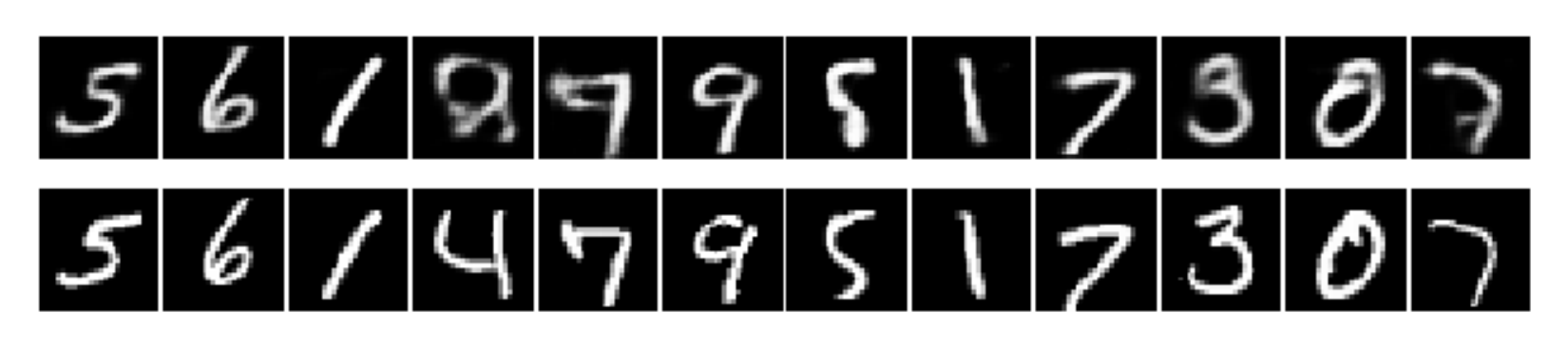} \\
        (a) GMMN MNIST samples & (b) GMMN TFD samples & (f) GMMN+AE nearest
        neighbors for MNIST samples \\
        \multirow{2}{*}[3.5em]{\includegraphics[width=0.25\textwidth]{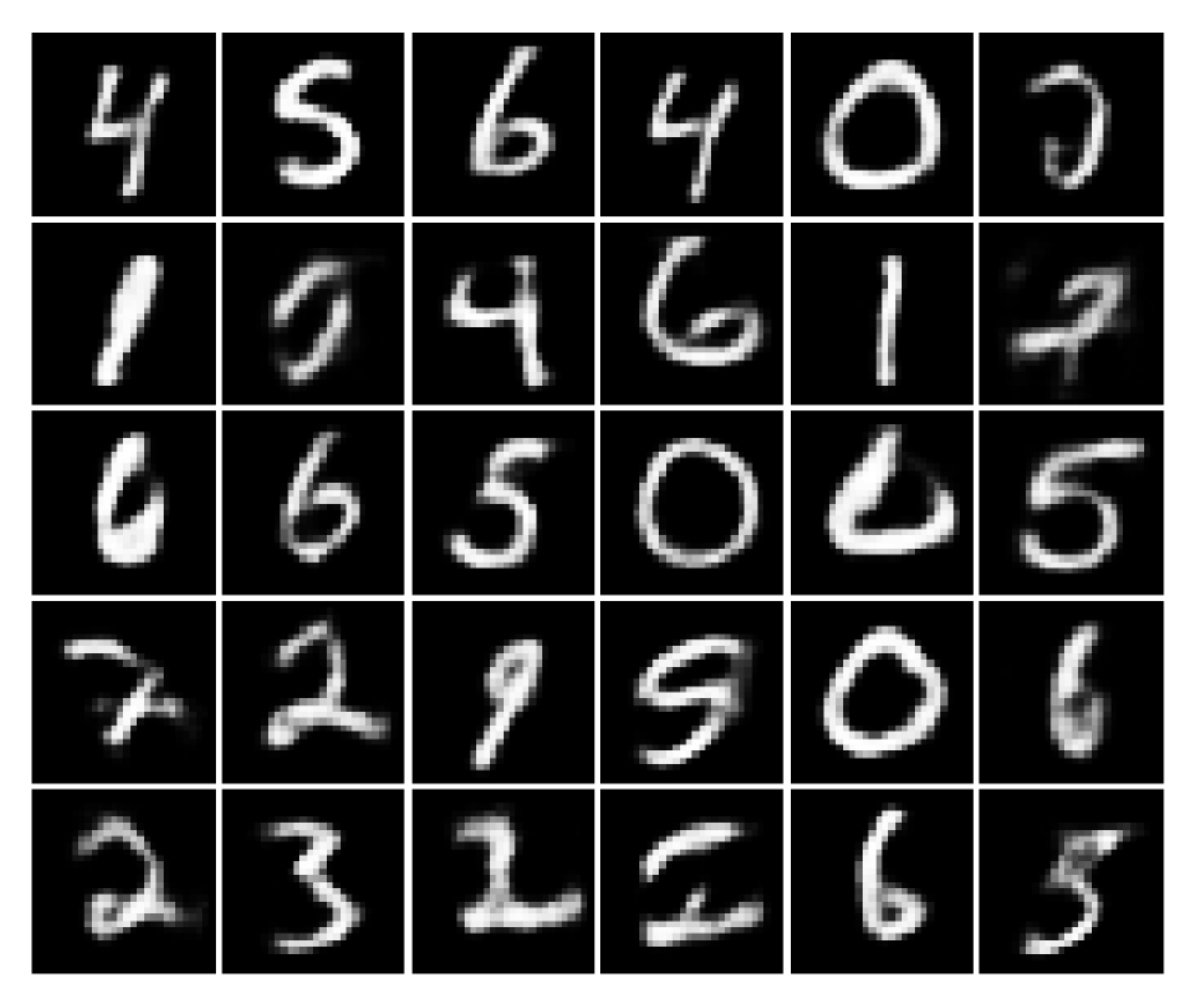}}
        &
        \multirow{2}{*}[3.5em]{\includegraphics[width=0.25\textwidth]{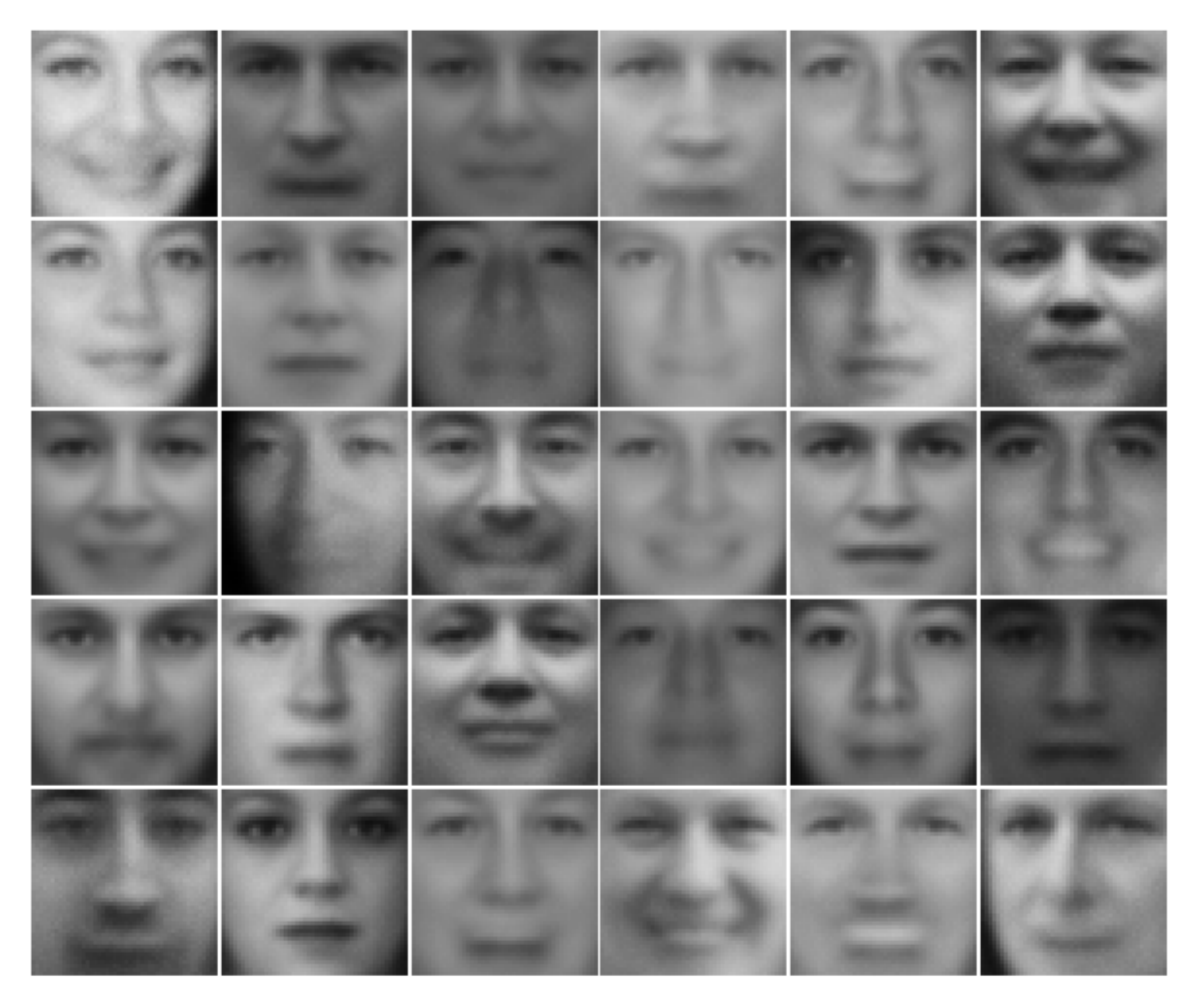}}
        &
        \includegraphics[width=0.45\textwidth]{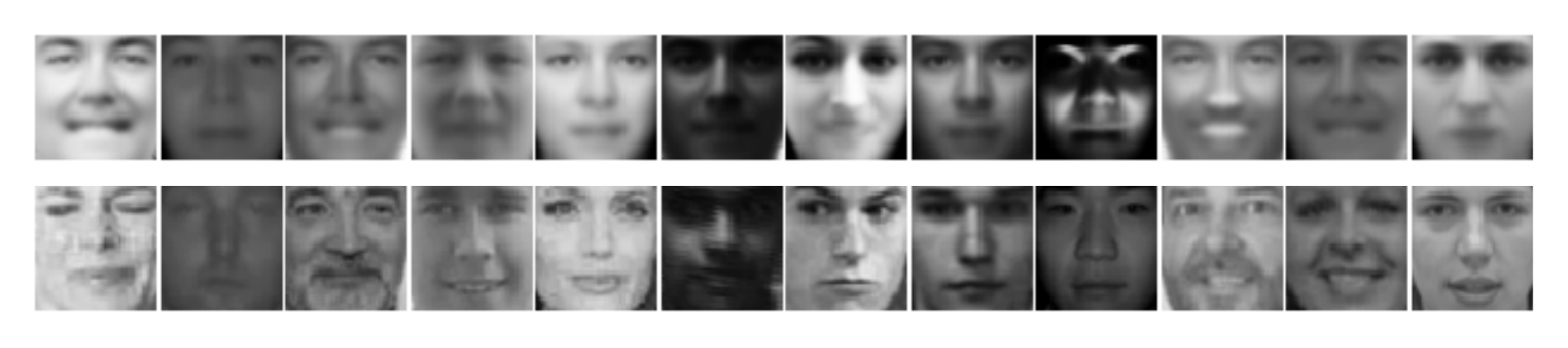} \\
        & & (g) GMMN nearest neighbors for TFD samples \\
        & & \includegraphics[width=0.45\textwidth]{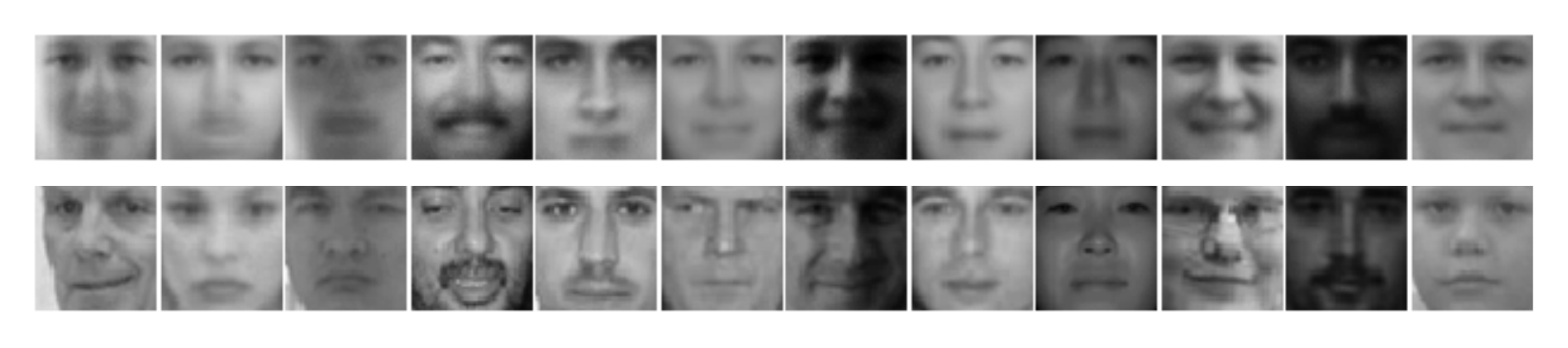} \\
        (c) GMMN+AE MNIST samples & (d) GMMN+AE TFD samples & (h) GMMN+AE nearest neighbors for TFD
        samples
    \end{tabular}
    \caption{Independent samples and their nearest neighbors in the training set for
    the GMMN+AE model trained on MNIST and TFD datasets. For (e)(f)(g) and (h) the
    top row are the samples from the model and the bottom row are the
    corresponding nearest neighbors from the training set measured by Euclidean
    distance.}
    \label{fig:samples}
\end{figure*}

We trained GMMNs on two benchmark datasets MNIST \cite{lecun1998gradient} and the
Toronto Face Dataset (TFD) \cite{susskind2010toronto}. For MNIST, we used the
standard test set of 10,000 images, and split out 5000 from the standard 60,000 training
images for validation. The remaining 55,000 were used for training. For TFD, we
used the same training and test sets and fold splits as used by
\cite{goodfellow2014generative}, but split out a small set of the training
data and used it as the validation set. For both datasets, rescaling the images to have pixel
intensities between 0 and 1 is the only preprocessing step we did.

On both datasets, we trained the GMMN network in both the input data space and the code space
of an auto-encoder. For all the networks we used in this section, a uniform distribution in $[-1,1]^H$ was used as the prior for the $H$-dimensional
stochastic hidden layer at the top of the GMMN, which was followed by 4
ReLU layers, and the output was a
layer of logistic sigmoid units.  The auto-encoder we used for MNIST had 4
layers, 2 for the encoder and 2 for the decoder.  For TFD the auto-encoder
had 6 layers in total, 3 for the encoder and 3 for the decoder. For both
auto-encoders the encoder and the decoder had mirrored
architectures.  All layers in the auto-encoder network used sigmoid
nonlinearities, which also guaranteed that the code space dimensions lay
in $[0,1]$, so that they could match the GMMN outputs. The network architectures for
MNIST are shown in Figure \ref{fig:architecture}.

The auto-encoders were trained separately from the GMMN. Cross entropy was used as the
reconstruction loss. We first did standard layer-wise pretraining, then
fine-tuned all layers jointly. Dropout \cite{hinton2012improving} was used on
the encoder layers. After training the auto-encoder, we fixed it and passed the
input data through the encoder to get the corresponding codes.
The GMMN network was then trained in this code space to match the statistics
of generated codes to the statistics of codes from data examples.
When generating samples, the generated codes
were passed through the decoder to get samples in the input data space.

For all experiments in this section the GMMN networks were trained with
minibatches of size 1000, for each minibatch we generated a set of 1000 samples from the network. The loss and gradient were computed from these
2000 points. We used the square root loss function $\mcL_\MMD$ throughout.

Evaluation of our model is not straight-forward, as we do not have an explicit
form for the probability density function, it is not easy to compute
the log-likelihood of data. However, sampling from our model is easy. We
therefore followed the same evaluation protocol used in related
models \cite{bengio2013better}, \cite{bengio2014deep}, and \cite{goodfellow2014generative}. A Gaussian Parzen window (kernel density
estimator) was fit to 10,000 samples generated from the model. The likelihood
of the test data was then computed under this distribution. The scale parameter
of the Gaussians was selected using a grid search in a fixed range using the
validation set.

The hyperparameters of the networks, including the learning rate and momentum
for both auto-encoder and GMMN training, dropout rate for the auto-encoder, and
number of hidden units on each layer of both auto-encoder and GMMN, were tuned
using Bayesian optimization~\cite{snoek-etal-2012,snoek-etal-2014}\footnote{We
used the service provided by \url{https://www.whetlab.com}} to optimize the
validation set likelihood under the Gaussian Parzen window density estimation. 

\begin{table}[t]
    \centering
    \begin{tabular}{c|c|c}
        \hline
        \hline
        Model & MNIST & TFD \\
        \hline
        DBN & 138 $\pm$ 2 & 1909 $\pm$ 66 \\
        Stacked CAE & 121 $\pm$ 1.6 & 2110 $\pm$ 50 \\
        Deep GSN & 214 $\pm$ 1.1 & 1890 $\pm$ 29 \\
        Adversarial nets & 225 $\pm$ 2 & 2057 $\pm$ 26 \\
        \hline
        GMMN & 147 $\pm$ 2 & 2085 $\pm$ 25 \\
     GMMN+AE & \bf{282 $\pm$ 2} & \bf{2204 $\pm$ 20} \\
        \hline
        \hline
    \end{tabular}
    \caption{Log-likelihood of the test sets under different models. The
    baselines are Deep Belief Net (DBN) and Stacked Contractive Auto-Encoder
    (Stacked CAE) from \cite{bengio2013better},
    Deep Generative Stochastic Network (Deep GSN) from \cite{bengio2014deep} and Adversarial nets (GANs) from
    \cite{goodfellow2014generative}.}
    \label{tbl:results}
    \vspace{-5pt}
\end{table}

The log-likelihood of the test set for both datasets are shown in Table
\ref{tbl:results}. The GMMN is competitive with other approaches, while
the GMMN+AE significantly outperforms the other models. This shows that despite being
relatively simple, MMD, especially when combined with an effective decoder,
is a powerful objective for training good generative models.

Some samples generated from the GMMN models are shown in Figure
\ref{fig:samples}(a-d). The GMMN+AE produces the most visually appealing samples,
which are reflected in its Parzen window log-likelihood estimates. The likely
explanation is that any perturbations in the code space correspond to smooth
transformations along the manifold of the data space. In that sense, the
decoder is able to ``correct'' noise in the code space.

To determine whether the models learned to merely copy the data, we follow the example
of~\cite{goodfellow2014generative} and visualize the nearest neighbour of
several samples in terms of Euclidean pixel-wise distance in Figure
\ref{fig:samples}(e-h). By this metric, it appears as though
the samples are not merely data examples.
\FloatBarrier

\begin{figure}[tb!]
    \centering
    \subfigure[MNIST interpolation]{
        \includegraphics[width=0.9\columnwidth]{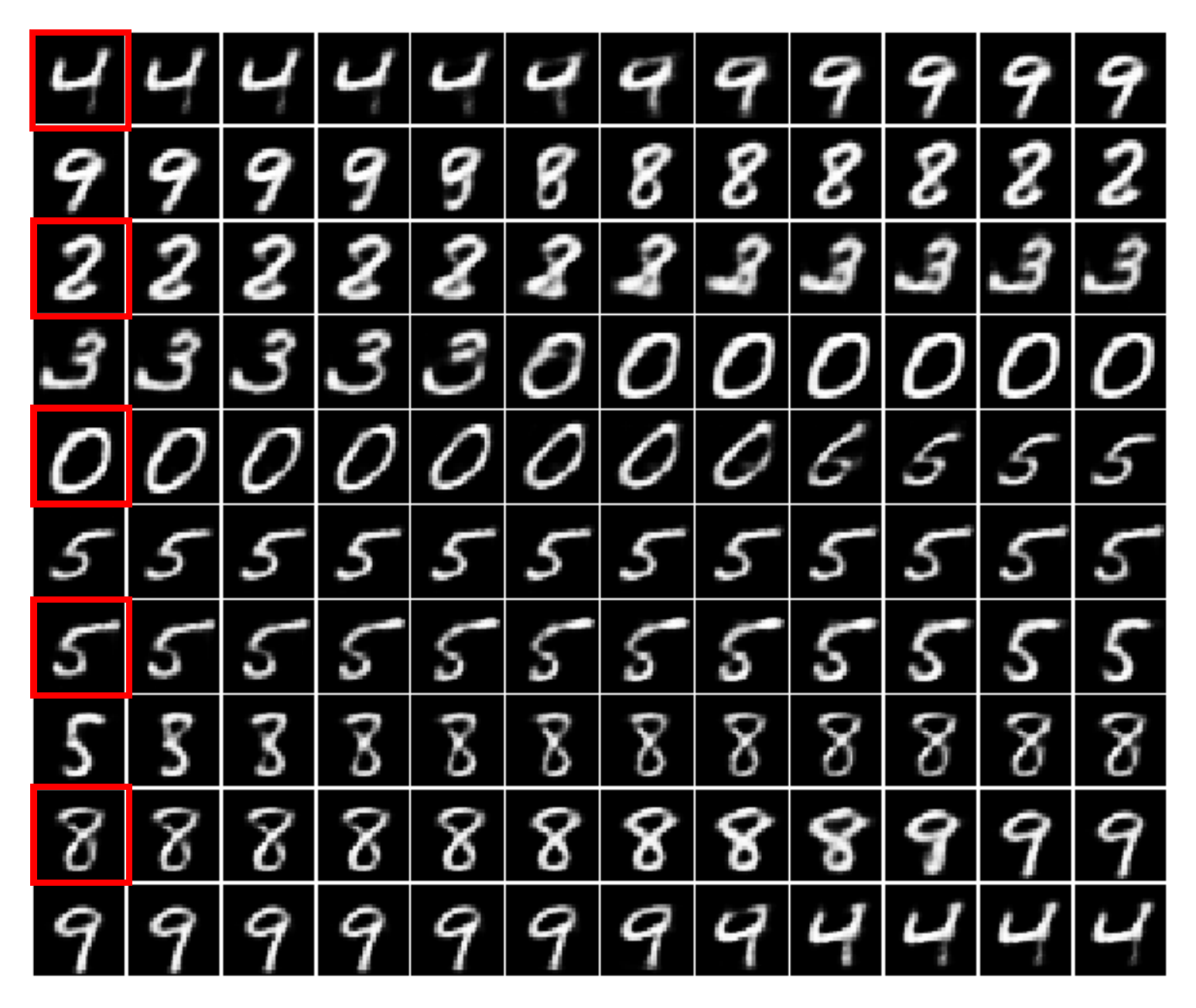}
        \label{subfig:mnistmorph}
    }

    \subfigure[TFD interpolation]{
        \includegraphics[width=0.9\columnwidth]{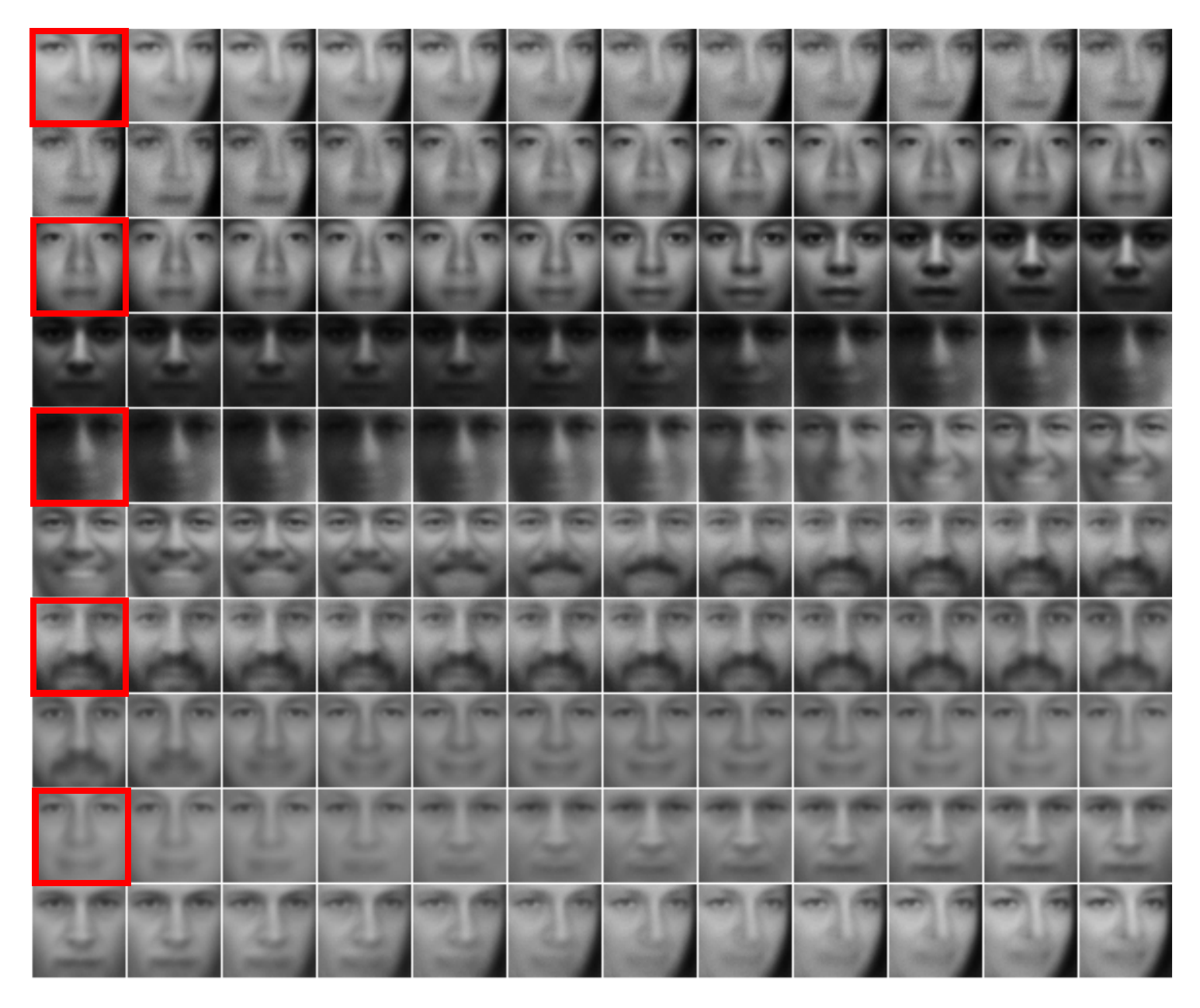}
        \label{subfig:tfdmorph}
    }
    \caption{Linear interpolation between 5 uniform random points from the GMMN+AE prior projected through the network into data space for \subref{subfig:mnistmorph} MNIST and \subref{subfig:tfdmorph} TFD. The 5 random points are highlighted with red boxes, and the interpolation goes from left to right, top to bottom. The final two rows represent an interpolation between the last highlighted image and the first highlighted image.}
    \label{fig:morph}
\end{figure}

One of the interesting aspects of a deep generative model such as the GMMN is
that it is possible to directly explore the data manifold. Using the GMMN+AE
model, we randomly sampled 5 points in the uniform space and show their
corresponding data space projections in Figure~\ref{fig:morph}.
These points are highlighted by red boxes. From left to right, top to bottom we linearly
interpolate between these points in the uniform space and show their corresponding
projections in data space. The manifold is smooth for the most part, and
almost all of the projections correspond to realistic looking data. For TFD in
particular, these transformations involve complex attributes, such as the
changing of pose, expression, lighting, gender, and facial hair.

\section{Conclusion and Future Work}
In this paper we provide a simple and effective framework for training deep
generative models called generative moment matching networks. Our approach is
based off of optimizing maximum mean discrepancy so that samples generated
from the model are indistinguishable from data examples in terms of their
moment statistics. As is standard with MMD, the use of the kernel trick allows
a GMMN to avoid explicitly computing these moments, resulting in a simple
training objective, and the use of minibatch stochastic gradient descent
allows the training to scale to large datasets.

Our second contribution combines MMD with auto-encoders for learning a
generative model of the code layer.
%in order to generate samples of the encoding layer.
The code samples from the model can then be fed through the
decoder in order to generate samples in the original space. 
%Unlike previous
%methods, the use of auto-encoders in this situation provides an elegant way to learn
%complex distributions with a relatively simple learning algorithm. 
The use of auto-encoders makes the generative model learning a much simpler
problem. Combined with MMD, pretrained auto-encoders can be readily bootstrapped
into a good generative model of data.
On the
MNIST and Toronto Face Database, the GMMN+AE model achieves superior
performance compared to other approaches. For these datasets, we demonstrate
that the GMMN+AE is able to discover the implicit manifold of the data.

There are many interesting directions for research using MMD. One such extension is to
consider alternatives to the standard MMD criterion in order to speed up training.
One such possibility is the class of linear-time estimators that has been
developed recently in the literature \cite{gretton2012kernel}.

Another possibility is to utilize random
features \cite{rahimi2007random}. These are randomized feature expansions whose inner product converges
to a kernel function with an increasing number of features. This idea was recently explored
for MMD in \cite{zhao2014fastmmd}. The advantage of this approach would be that the cost would no longer
grow quadratically with minibatch size because we could use the original objective given in Equation \ref{eq:mmdprimal}. Another advantage of this approach is that the data statistics could be pre-computed from the entire dataset, which would reduce the variance of the objective gradients.

Another direction we would like to explore is joint training of the auto-encoder model with the GMMN. Currently, these are treated separately, but joint training may encourage the learning of codes that are both suitable for reconstruction as well as generation.

While a GMMN provides an easy way to sample data, the posterior distribution
over the latent variables is not readily available. It would be interesting to
explore ways in which to infer the posterior distribution over the latent
space. A straightforward way to do this is to learn a neural network to
predict the latent vector given a sample. This is reminiscent of the
recognition models used in the wake-sleep algorithm \cite{hinton1995wake}, or
variational auto-encoders \cite{kingma2014variational}.

An interesting application of MMD that is not directly related to generative
modelling comes from recent work on learning fair representations~\cite{zemel2013learning}.
There, the objective is to train a prediction method that is invariant to a
particular sensitive attribute of the data. Their solution is to learn an
intermediate clustering-based representation. MMD could instead be applied
to learn a more powerful, distributed representation such that the statistics
of the representation do not change conditioned on the sensitive variable.
This idea can be further generalized to learn representations invariant to
known biases.

Finally, the notion of utilizing an auto-encoder with the GMMN+AE model provides new avenues for creating generative models of even more complex datasets. For example, it may
be possible to use a GMMN+AE with convolutional auto-encoders
\cite{zeiler2010deconvolutional,masci2011stacked,makhzaniconvolutional2014} in order to create generative models of high resolution color images.

\section*{Acknowledgements}
We thank David Warde-Farley for helpful clarifications
regarding~\cite{goodfellow2014generative}, and Charlie Tang for providing
relevant references. We thank CIFAR, NSERC, and Google for research funding.

%\section{Conclusion}
%\input{conclusion}

\bibliography{generativenets}

\begin{thebibliography}{46}
\providecommand{\natexlab}[1]{#1}
\providecommand{\url}[1]{\texttt{#1}}
\expandafter\ifx\csname urlstyle\endcsname\relax
  \providecommand{\doi}[1]{doi: #1}\else
  \providecommand{\doi}{doi: \begingroup \urlstyle{rm}\Url}\fi

\bibitem[Ackley et~al.(1985)Ackley, Hinton, and Sejnowski]{ackley1985learning}
Ackley, D.~H., Hinton, G.~E., and Sejnowski, T.~J.
\newblock A learning algorithm for boltzmann machines.
\newblock \emph{Cognitive science}, 9\penalty0 (1):\penalty0 147--169, 1985.

\bibitem[Bengio et~al.(2007)Bengio, Lamblin, Popovici, Larochelle,
  et~al.]{bengio2007greedy}
Bengio, Y., Lamblin, P., Popovici, D., Larochelle, H., et~al.
\newblock Greedy layer-wise training of deep networks.
\newblock In \emph{Advances in Neural Information Processing Systems (NIPS)},
  2007.

\bibitem[Bengio et~al.(2013{\natexlab{a}})Bengio, Mesnil, Dauphin, and
  Rifai]{bengio2013better}
Bengio, Y., Mesnil, G., Dauphin, Y., and Rifai, S.
\newblock Better mixing via deep representations.
\newblock In \emph{Proceedings of the 28th International Conference on Machine
  Learning (ICML)}, 2013{\natexlab{a}}.

\bibitem[Bengio et~al.(2013{\natexlab{b}})Bengio, Yao, Alain, and
  Vincent]{bengio2013generalized}
Bengio, Y., Yao, L., Alain, G., and Vincent, P.
\newblock Generalized denoising auto-encoders as generative models.
\newblock In \emph{Advances in Neural Information Processing Systems}, pp.\
  899--907, 2013{\natexlab{b}}.

\bibitem[Bengio et~al.(2014)Bengio, Thibodeau-Laufer, Alain, and
  Yosinski]{bengio2014deep}
Bengio, Y., Thibodeau-Laufer, E., Alain, G., and Yosinski, J.
\newblock Deep generative stochastic networks trainable by backprop.
\newblock In \emph{Proceedings of the 29th International Conference on Machine
  Learning (ICML)}, 2014.

\bibitem[Cho et~al.(2014)Cho, van Merrienboer, Gulcehre, Bougares, Schwenk, and
  Bengio]{cho2014learning}
Cho, K., van Merrienboer, B., Gulcehre, C., Bougares, F., Schwenk, H., and
  Bengio, Y.
\newblock Learning phrase representations using rnn encoder-decoder for
  statistical machine translation.
\newblock In \emph{Conference on Empirical Methods in Natural Language
  Processing (EMNLP)}, 2014.

\bibitem[Fang et~al.(2014)Fang, Gupta, Iandola, Srivastava, Deng, Doll{\'a}r,
  Gao, He, Mitchell, Platt, Zitnick, and Zweig]{fang2014captions}
Fang, H., Gupta, S., Iandola, F., Srivastava, R., Deng, L., Doll{\'a}r, P.,
  Gao, J., He, X., Mitchell, M., Platt, J., Zitnick, C.~L., and Zweig, G.
\newblock From captions to visual concepts and back.
\newblock \emph{arXiv preprint arXiv:1411.4952}, 2014.

\bibitem[Goodfellow et~al.(2014)Goodfellow, Pouget-Abadie, Mirza, Xu,
  Warde-Farley, Ozair, Courville, and Bengio]{goodfellow2014generative}
Goodfellow, I., Pouget-Abadie, J., Mirza, M., Xu, B., Warde-Farley, D., Ozair,
  S., Courville, A., and Bengio, Y.
\newblock Generative adversarial nets.
\newblock In \emph{Advances in Neural Information Processing Systems}, pp.\
  2672--2680, 2014.

\bibitem[Graves \& Jaitly(2014)Graves and Jaitly]{graves2014towards}
Graves, A. and Jaitly, N.
\newblock Towards end-to-end speech recognition with recurrent neural networks.
\newblock In \emph{Proceedings of the 31st International Conference on Machine
  Learning (ICML-14)}, pp.\  1764--1772, 2014.

\bibitem[Gretton et~al.(2007)Gretton, Borgwardt, Rasch, Sch{\"o}lkopf, and
  Smola]{gretton2007kernel}
Gretton, A., Borgwardt, K.~M., Rasch, M., Sch{\"o}lkopf, B., and Smola, A.~J.
\newblock A kernel method for the two-sample-problem.
\newblock In \emph{Advances in Neural Information Processing Systems (NIPS)},
  2007.

\bibitem[Gretton et~al.(2012{\natexlab{a}})Gretton, Borgwardt, Rasch,
  Sch{\"o}lkopf, and Smola]{gretton2012kernel}
Gretton, A., Borgwardt, K.~M., Rasch, M.~J., Sch{\"o}lkopf, B., and Smola, A.
\newblock A kernel two-sample test.
\newblock \emph{The Journal of Machine Learning Research}, 13\penalty0
  (1):\penalty0 723--773, 2012{\natexlab{a}}.

\bibitem[Gretton et~al.(2012{\natexlab{b}})Gretton, Sejdinovic, Strathmann,
  Balakrishnan, Pontil, Fukumizu, and Sriperumbudur]{gretton2012optimal}
Gretton, A., Sejdinovic, D., Strathmann, H., Balakrishnan, S., Pontil, M.,
  Fukumizu, K., and Sriperumbudur, B.~K.
\newblock Optimal kernel choice for large-scale two-sample tests.
\newblock In \emph{Advances in Neural Information Processing Systems}, pp.\
  1205--1213, 2012{\natexlab{b}}.

\bibitem[Hinton(2002)]{hinton2002training}
Hinton, G.~E.
\newblock Training products of experts by minimizing contrastive divergence.
\newblock \emph{Neural Computation}, 14\penalty0 (8):\penalty0 1771--1800,
  2002.

\bibitem[Hinton et~al.(1995)Hinton, Dayan, Frey, and Neal]{hinton1995wake}
Hinton, G.~E., Dayan, P., Frey, B.~J., and Neal, R.~M.
\newblock The ``wake-sleep'' algorithm for unsupervised neural networks.
\newblock \emph{Science}, 268\penalty0 (5214):\penalty0 1158--1161, 1995.

\bibitem[Hinton et~al.(2012{\natexlab{a}})Hinton, Deng, Yu, Dahl, Mohamed,
  Jaitly, Senior, Vanhoucke, Nguyen, Sainath, and
  Kingsbury]{deepSpeechReviewSPM2012}
Hinton, G.~E., Deng, L., Yu, D., Dahl, G.~E., Mohamed, A., Jaitly, N., Senior,
  A., Vanhoucke, V., Nguyen, P., Sainath, T.~N., and Kingsbury, B.
\newblock Deep neural networks for acoustic modeling in speech recognition: The
  shared views of four research groups.
\newblock \emph{IEEE Signal Process. Mag.}, 29\penalty0 (6):\penalty0 82--97,
  2012{\natexlab{a}}.

\bibitem[Hinton et~al.(2012{\natexlab{b}})Hinton, Srivastava, Krizhevsky,
  Sutskever, and Salakhutdinov]{hinton2012improving}
Hinton, G.~E., Srivastava, N., Krizhevsky, A., Sutskever, I., and
  Salakhutdinov, R.~R.
\newblock Improving neural networks by preventing co-adaptation of feature
  detectors.
\newblock \emph{arXiv preprint arXiv:1207.0580}, 2012{\natexlab{b}}.

\bibitem[Kingma \& Welling(2014)Kingma and Welling]{kingma2014variational}
Kingma, D.~P. and Welling, M.
\newblock Auto-encoding variational {B}ayes.
\newblock In \emph{International Conference on Learning Representations}, 2014.

\bibitem[Kiros et~al.(2014)Kiros, Salakhutdinov, and Zemel]{kiros2014unifying}
Kiros, R., Salakhutdinov, R., and Zemel, R.~S.
\newblock Unifying visual-semantic embeddings with multimodal neural language
  models.
\newblock \emph{arXiv preprint arXiv:1411.2539}, 2014.

\bibitem[Krizhevsky et~al.(2012)Krizhevsky, Sutskever, and
  Hinton]{krizhevsky2012imagenet}
Krizhevsky, A., Sutskever, I., and Hinton, G.~E.
\newblock Imagenet classification with deep convolutional neural networks.
\newblock In \emph{Advances in Neural Information Processing Systems (NIPS)},
  2012.

\bibitem[Larochelle \& Murray(2011)Larochelle and Murray]{larochelle2011neural}
Larochelle, H. and Murray, I.
\newblock The neural autoregressive distribution estimator.
\newblock In \emph{roceedings of the 14th International Conference on
  Artificial Intelligence and Statistics (AISTATS)}, 2011.

\bibitem[LeCun et~al.(1998)LeCun, Bottou, Bengio, and
  Haffner]{lecun1998gradient}
LeCun, Y., Bottou, L., Bengio, Y., and Haffner, P.
\newblock Gradient-based learning applied to document recognition.
\newblock \emph{Proceedings of the IEEE}, 86\penalty0 (11):\penalty0
  2278--2324, 1998.

\bibitem[MacKay(1995)]{mackay1995bayesian}
MacKay, D.~J.
\newblock Bayesian neural networks and density networks.
\newblock \emph{Nuclear Instruments and Methods in Physics Research Section A:
  Accelerators, Spectrometers, Detectors and Associated Equipment},
  354\penalty0 (1):\penalty0 73--80, 1995.

\bibitem[Magdon-Ismail \& Atiya(1998)Magdon-Ismail and Atiya]{magdon1998neural}
Magdon-Ismail, M. and Atiya, A.
\newblock Neural networks for density estimation.
\newblock In \emph{NIPS}, pp.\  522--528, 1998.

\bibitem[Makhzani \& Frey(2014)Makhzani and Frey]{makhzaniconvolutional2014}
Makhzani, A. and Frey, B.
\newblock A winner-take-all method for training sparse convolutional
  autoencoders.
\newblock In \emph{NIPS Deep Learning Workshop}, 2014.

\bibitem[Masci et~al.(2011)Masci, Meier, Cire{\c{s}}an, and
  Schmidhuber]{masci2011stacked}
Masci, J., Meier, U., Cire{\c{s}}an, D., and Schmidhuber, J.
\newblock Stacked convolutional auto-encoders for hierarchical feature
  extraction.
\newblock In \emph{Artificial Neural Networks and Machine Learning--ICANN
  2011}, pp.\  52--59. Springer, 2011.

\bibitem[Mnih \& Gregor(2014)Mnih and Gregor]{mnih2014belief}
Mnih, A. and Gregor, K.
\newblock Neural variational inference and learning in belief networks.
\newblock In \emph{International Conference on Machine Learning}, 2014.

\bibitem[Nair \& Hinton(2010)Nair and Hinton]{nair2010rectified}
Nair, V. and Hinton, G.~E.
\newblock Rectified linear units improve restricted boltzmann machines.
\newblock In \emph{International Conference on Machine Learning}, pp.\
  807--814, 2010.

\bibitem[Neal(1992)]{neal1992connectionist}
Neal, R.~M.
\newblock Connectionist learning of belief networks.
\newblock \emph{Artificial intelligence}, 56\penalty0 (1):\penalty0 71--113,
  1992.

\bibitem[Rahimi \& Recht(2007)Rahimi and Recht]{rahimi2007random}
Rahimi, A. and Recht, B.
\newblock Random features for large-scale kernel machines.
\newblock In \emph{Advances in Neural Information Processing Systems (NIPS)},
  2007.

\bibitem[Ramdas et~al.(2015)Ramdas, Reddi, Poczos, Singh, and
  Wasserman]{ramadas2015decreasing}
Ramdas, A., Reddi, S.~J., Poczos, B., Singh, A., and Wasserman, L.
\newblock On the decreasing power of kernel and distance based nonparametric
  hypothesis tests in high dimensions.
\newblock In \emph{The Twenty-Ninth AAAI Conference on Artificial Intelligence
  (AAAI-15)}, 2015.

\bibitem[Rezende et~al.(2014)Rezende, Mohamed, and
  Wierstra]{rezende2014stochastic}
Rezende, D.~J., Mohamed, S., and Wierstra, D.
\newblock Stochastic backpropagation and approximate inference in deep
  generative models.
\newblock In \emph{International Conference on Machine Learning}, pp.\
  1278--1286, 2014.

\bibitem[Rifai et~al.(2011)Rifai, Vincent, Muller, Glorot, and
  Bengio]{rifai2011contractive}
Rifai, S., Vincent, P., Muller, X., Glorot, X., and Bengio, Y.
\newblock Contractive auto-encoders: Explicit invariance during feature
  extraction.
\newblock In \emph{Proceedings of the 28th International Conference on Machine
  Learning (ICML-11)}, pp.\  833--840, 2011.

\bibitem[Rifai et~al.(2012)Rifai, Bengio, Dauphin, and
  Vincent]{rifai2012generative}
Rifai, S., Bengio, Y., Dauphin, Y., and Vincent, P.
\newblock A generative process for sampling contractive auto-encoders.
\newblock In \emph{International Conference on Machine Learning (ICML)}, 2012.

\bibitem[Salakhutdinov \& Hinton(2009)Salakhutdinov and
  Hinton]{salakhutdinov2009deep}
Salakhutdinov, R. and Hinton, G.~E.
\newblock Deep boltzmann machines.
\newblock In \emph{International Conference on Artificial Intelligence and
  Statistics}, 2009.

\bibitem[Sermanet et~al.(2014)Sermanet, Eigen, Zhang, Mathieu, Fergus, and
  LeCun]{sermanet2014overfeat}
Sermanet, P., Eigen, D., Zhang, X., Mathieu, M., Fergus, R., and LeCun, Y.
\newblock Overfeat: Integrated recognition, localization and detection using
  convolutional networks.
\newblock In \emph{International Conference on Learning Representations}, 2014.

\bibitem[Snoek et~al.(2012)Snoek, Larochelle, and Adams]{snoek-etal-2012}
Snoek, J., Larochelle, H., and Adams, R.~P.
\newblock Practical {B}ayesian optimization of machine learning algorithms.
\newblock In \emph{Advances in Neural Information Processing Systems}, 2012.

\bibitem[Snoek et~al.(2014)Snoek, Swersky, Zemel, and Adams]{snoek-etal-2014}
Snoek, J., Swersky, K., Zemel, R.~S., and Adams, R.~P.
\newblock Input warping for bayesian optimization of non-stationary functions.
\newblock In \emph{International Conference on Machine Learning}, 2014.

\bibitem[Sriperumbudur et~al.(2009)Sriperumbudur, Fukumizu, Gretton, Lanckriet,
  and Sch{\"o}lkopf]{sriperumbudur2009kernel}
Sriperumbudur, B.~K., Fukumizu, K., Gretton, A., Lanckriet, G.~R., and
  Sch{\"o}lkopf, B.
\newblock Kernel choice and classifiability for rkhs embeddings of probability
  distributions.
\newblock In \emph{Advances in Neural Information Processing Systems}, pp.\
  1750--1758, 2009.

\bibitem[Susskind et~al.(2010)Susskind, Anderson, and
  Hinton]{susskind2010toronto}
Susskind, J., Anderson, A., and Hinton, G.~E.
\newblock The toronto face dataset.
\newblock Technical report, Department of Computer Science, University of
  Toronto, 2010.

\bibitem[Sutskever et~al.(2014)Sutskever, Vinyals, and
  Le]{sutskever2014sequence}
Sutskever, I., Vinyals, O., and Le, Q.~V.
\newblock Sequence to sequence learning with neural networks.
\newblock In \emph{Advances in Neural Information Processing Systems}, pp.\
  3104--3112, 2014.

\bibitem[Szegedy et~al.(2014)Szegedy, Liu, Jia, Sermanet, Reed, Anguelov,
  Erhan, Vanhoucke, and Rabinovich]{szegedy2014going}
Szegedy, C., Liu, W., Jia, Y., Sermanet, P., Reed, S., Anguelov, D., Erhan, D.,
  Vanhoucke, V., and Rabinovich, A.
\newblock Going deeper with convolutions.
\newblock \emph{arXiv preprint arXiv:1409.4842}, 2014.

\bibitem[Vincent et~al.(2008)Vincent, Larochelle, Bengio, and
  Manzagol]{vincent2008extracting}
Vincent, P., Larochelle, H., Bengio, Y., and Manzagol, P.-A.
\newblock Extracting and composing robust features with denoising autoencoders.
\newblock In \emph{Proceedings of the 25th international conference on Machine
  learning}, pp.\  1096--1103. ACM, 2008.

\bibitem[Vinyals et~al.(2014)Vinyals, Toshev, Bengio, and
  Erhan]{vinyals2014show}
Vinyals, O., Toshev, A., Bengio, S., and Erhan, D.
\newblock Show and tell: A neural image caption generator.
\newblock \emph{arXiv preprint arXiv:1411.4555}, 2014.

\bibitem[Zeiler et~al.(2010)Zeiler, Krishnan, Taylor, and
  Fergus]{zeiler2010deconvolutional}
Zeiler, M.~D., Krishnan, D., Taylor, G.~W., and Fergus, R.
\newblock Deconvolutional networks.
\newblock In \emph{Computer Vision and Pattern Recognition}, pp.\  2528--2535.
  IEEE, 2010.

\bibitem[Zemel et~al.(2013)Zemel, Wu, Swersky, Pitassi, and
  Dwork]{zemel2013learning}
Zemel, R., Wu, Y., Swersky, K., Pitassi, T., and Dwork, C.
\newblock Learning fair representations.
\newblock In \emph{International Conference on Machine Learning}, pp.\
  325--333, 2013.

\bibitem[Zhao \& Meng(2014)Zhao and Meng]{zhao2014fastmmd}
Zhao, J. and Meng, D.
\newblock Fastmmd: Ensemble of circular discrepancy for efficient two-sample
  test.
\newblock \emph{arXiv preprint arXiv:1405.2664}, 2014.

\end{thebibliography}
\bibliographystyle{icml2015}

\end{document}